# The First Parallel Multilingual Corpus of Persian: Toward a Persian BLARK

Behrang Qasemizadeh[1], Saeed Rahimi[2], Behrooz Mahmoodi Bakhtiari[3]

[1] Researcher, Text and Speech Technology LTD., qasemizadeh@comp.iust.ac.ir
[2] Graduate Student, University of Tehran and TST LTD., saeedrahimiavval@yahoo.com
[3] Assistant Professor, University of Tehran, b_m_bakhtiari@yahoo.com

## Abstract

In this article, we have introduced the first parallel corpus of Persian with more than 10 other European languages. This article describes primary steps toward preparing a Basic Language Resources Kit (BLARK) for Persian. Up to now, we have proposed morphosyntactic specification of Persian based on EAGLE/MULTEXT guidelines and specific resources of MULTEXT-East. The article introduces Persian Language, with emphasis on its orthography and morphosyntactic features, then a new Part-of-Speech categorization and orthography for Persian in digital environments is proposed. Finally, the corpus and related statistic will be analyzed.

## 1- Introduction

With the expansion of information and communication technology, the need for HLT also increases. In order for people to use their native language in these applications, a set of basic provisions (such as tools, corpora, and lexicons) is required. Since now, there have been lots of efforts toward preparing Basic Language Resources Kit for language processing, particularly languages that are commercially less considered. The outcomes of such efforts are corpora, morphological analyzer etc., which are freely available for train and research purposes. [1] As to Persian, or Farsi, unfortunately there has been little work in this area, or at least these efforts have not reached that ultimate goal, preparing Persian BLARK. [2] Moreover, these works have not been done based on certain specific frameworks or standards which can be extended. Last but not the least, is that even these resources are not easily accessible for research or training programs.

In our previous work [3], we reported the first steps towards preparing a BLARK for Persian. This comprises of an introduction to the orthography of Persian in digital media (A convention for representing Persian e-text, especially for Persian Corpora) and its encoding, introducing Persian Part of Speeches (PoS), morphosyntactic specification of Persian according to EAGLE/MULTEXT guidelines and proposed standards in MULTEXT-East.

The MULTEXT-East project was a branch of the EU MULTEXT [5] project. It developed standardized language resources for six languages such as Bulgarian, Czech, Estonian, Hungarian, Romanian, Slovene, as well as English, the 'hub' language of the project. The main result of the project was an annotated multilingual corpus, comprising a speech corpus, a comparable corpus and a parallel corpus, together with lexical resources, and tool resources for these seven languages. The most useful part of the MULTEXT-East project was the morphosyntactic resources which consist of three layers, listed in order of abstraction as follows:

- 1. *1984* MSD: the morphosyntactically annotated Orwell's 1984 corpus, in which each word is assigned its context-disambiguated MSD and lemma.
- 2. MSD Lexicons: the morphosyntactic lexicons, which contain the full inflectional paradigms of a superset of the lemmas that appear in the 1984 corpus. Each entry gives the word-form, its lemma and MSD.
- 3. MSD Specs: the morphosyntactic specifications, which set out the grammar of the valid morphosyntactic descriptions, MSDs. The specifications determine what valid MSD for each language is, and what it may mean, e.g., "Ncms" means PoS: Noun, Type: common, Gender: masculine, Number: singular.

In this way, MULTEXT-East provides a comprehensive framework for corpus development. Also, there are lots of resources according to this framework, e.g. '1984 MSD' for several languages. On the other hand, 1984 is available in Persian and in return Persian as an Indo-European language, can be placed with this framework as we showed in our previous work. In our previous work, we discussed related issues like Persian orthography and its encoding in digital media (e-orthography), the morphosyntactic specifications of Persian, and finally interaction of these elements and the effects of these definitions in the process of computational analysis of the language.

This article introduces the next steps toward this goal in a real application, preparing a corpus of Persian in MULTEXT-East framework parallel to the other languages which are available in this framework. The article is organized as follows: according to the importance of Persian orthography, first we will describe our previous work briefly. Then, the modified morphosyntactic specification of Persian and related tables are proposed. In the fourth section, we report statistic and related issues of the corpus. Finally, section five provides the conclusion, and discussion of our future programs.

## 2- Persian in MULTEXT-East Framework

As it was mentioned before, in our previous work we fitted Persian in MULTEXT-East framework and we proposed its morphosyntactic specifications according to that framework. In some cases, in keeping with especial morphosyntactic specifications of Persian, some features have been added to it. Adding Persian to this framework came along with some challenges; the most important of which were as follows:

- Issues related to encoding and representation of Persian e-texts, and at the same time, its orthography
- Provision of morphosyntactic specification of Persian according to real facts about the language and simultaneously its application in computational linguistics, and HLT
- Interaction of definitions in the first and second case, and simultaneously its effects on other steps in computational analysis of language such as tokenization (Fig. 1)

In order to solve the first problem we decided to propose an approach for Persian e-texts representation (let's say e-orthography) which is a combination of proposed standard by Iranian Standard Institute [6] and official orthography of Persian for the paper based system [7] by the Iranian Academy Of Persian Language and Literature. According to this proposed approach, the control character Zero With non-Joiner (ZWNJ) appears whereas semi-space appears in the paper-based system of writing. Therefore, all suffixes and prefixes, where they have to be written in separate form in article based system, are appeared with a ZWNJ character in electronic format. More comprehensive descriptions about this issue can be found in [3], and [8].

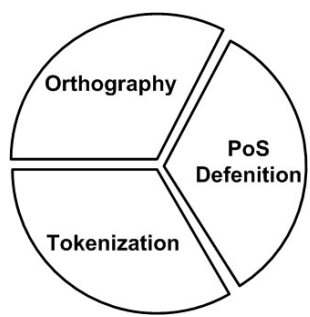

**Fig. 1.** Consider the whole circle as a proposed standard for corpus tagging in a specific language. Then the tokenization policy, PoS categorization, and language orthography, are fundamental elements that will directly affect the set of tags which is defined for corpus tagging.

As to the second issue, PoSs of Persian, according to our survey on Persian [9][10][11][12][13], we proposed a new PoS categorization for Persian in MULTEXT-East framework. This new categorization comprises of 11 different PoSs for Persian. Table 1 shows these PoSs, the number of attributes, and abbreviations for each PoS. Some of attributes are added to the framework to show the specific features of Persian which are not applicable in other languages; while some other attributes are not applicable to Persian.

According to the experimental results such as [14], the authors believe that proposed approach for Persian e-orthography and the new definitions for Persian MSDs can result in better outcome in computational analysis of this language. In the next section, we discuss the morphosyntactic specifications of Persian and its encoding in MULTEXT-East framework.

**Table 1. Persian PoSs and the number of attributes for each of them**

| Part of Speech | Code | Number of Attributes |
|---:|:---:|:---:|
| Noun | N | 4 |
| Verb | V | 10 |
| Adjective | A | 4 |
| Pronoun | P | 6 |
| Determiner | D | 1 |
| Adverb | R | 2 |
| Adposition | S | 2 |
| Conjunction | C | 2 |

| | | | |
|---|---|---|---|
| Numeral | M | 3 | |
| Interjection | I | 0 | |
| Abbreviation | Y | 0 | |

## 3- Persian Morphosyntactic Specification

MULTEXT-East framework uses a position based tagging system, and each specification is shown using one character. In this framework, the first character in each annotation shows the PoS of the word; then the value of attributes are indicated as characters which appear after the PoS character in predefined positions. In case that an attribute is not applicable to a certain language, a dash appears in the place of this attribute. More information about representing the attributes and morphosyntactic annotations of words may be found in [15].

In the following tables, morphosyntactic specifications of Persian for each PoS are displayed. In these tables, the first column on the left shows the position of value for each attribute in an annotation. To be concise, the attributes which are not applicable for Persian are omitted from the tables; for this reason, there is no value for some places in the tables. The second column in each table shows the name of attributes e.g. the 'Number' and 'Person' attribute for Verbs. The third and forth columns show meaningful values for each attribute and their abbreviation for them respectively.

As an example, for the 'Person' attribute of Verb PoS, there are 3 meaningful values which are *first, second* and *third,* that are shown by characters 1, 2, and 3 in the forth position after PoS character (V).

The tables 2 to 10 show morphosyntactic specification of PoSs of Persian, the encoding, and their values. These tables are not shown for Interjections and Abbreviations since they do not have any attributes.

**Table 2. Morphosyntactic specification of Nouns**

| P | ATT | VAL | C |
|---|---|---|---|
| 1 | Type | common | c |
| | | proper | p |
| 3 | Number | singular | s |
| | | plural | p |
| 4 | Case | genitive | g |
| | | vocative | v |
| 5 | Definiteness | no[1] | n |
| | | yes | y |

**Table 3. Morphosyntactic specification of Verbs**

| P | ATT | VAL | C |
|---|---|---|---|
| 1 | Type | main | m |
| | | auxiliary | a |
| | | modal | o |
| | | copula | c |
| | | light | l |
| 2 | VForm | indicative | i |
| | | subjunctive | s |
| | | imperative | m |
| | | participle | p |
| 3 | Tense | present | p |
| | | past | s[2] |
| 4 | Person | first | 1 |
| | | second | 2 |
| | | third | 3 |
| 5 | Number | singular | s |
| | | plural | p |
| 8 | Negative | no | n |
| | | yes | y |
| 10 | Clitic | no | n |
| | | yes | y |
| 14 | Aspect | progressive | p |
| 15 | Courtesy | no | n |
| | | yes | y |
| 16 | Transitive | no | n |
| | | yes | y |

**Table 4. Morphosyntactic specification of Adjectives**

| P | ATT | VAL | C |
|---|---|---|---|
| 1 | Type | qualificative | f |

---

[1] "no" indicates that a feature is absent and "yes" indicates the presence of a feature.

[2] Future tense was not considered, due to its periphrastic morphological shape.

```
- -------------- -------------- -
2 Degree         positive       p
                 comparative    c
                 superlative    s
- -------------- -------------- -
5 Case           genitive       g
- -------------- -------------- -
6 Definiteness   no             n
                 yes            y
================================
```

**Table 5. Morphosyntactic specification of Pronouns**
```
= ============== ============== =
P ATT            VAL            C
= ============== ============== =
1 Type           personal       p
                 demonstrative  d
                 indefinite     i
                 interrogative  q
                 reflexive      x
                 reciprocal     y
- -------------- -------------- -
2 Person         first          1
                 second         2
                 third          3
- -------------- -------------- -
4 Number         singular       s
                 plural         p
- -------------- -------------- -
5 Case           genitive       g
                 accusative     a
- -------------- -------------- -
8 Clitic         no             n
                 yes            y
- -------------- -------------- -
================================
```

**Table 6. Morphosyntactic specification of Determiner**
```
= ============== ============== =
P ATT            VAL            C
= ============== ============== =
1 Type           demonstrative  d
                 indefinite     i
                 interrogative  q
                 exclamative    e
                 article        a
                 exceptional³   x
- -------------- -------------- -
4 Number         singular       s
                 plural         p
- -------------- -------------- -
```

---

³ We use this term to refer the only member of this class: the word "tanhâ" (the only).

```
================================
```

**Table 7. Morphosyntactic specification of Adverbs**
```
= ============== ============== =
P ATT            VAL            C
= ============== ============== =
2 Degree         positive       p
                 comparative    c
- -------------- -------------- -
7 Case           genitive       g
================================
```

**Table 8. Morphosyntactic specification of Adpositions**
```
= ============== ============== =
P ATT            VAL            C
= ============== ============== =
1 Type           preposition    p
                 postposition   t
- -------------- -------------- -
2 Formation      simple         s
                 compound       c
================================
```

**Table 9. Morphosyntactic specification of Conjunctions**
```
= ============== ============== =
P ATT            VAL            C
= ============== ============== =
1 Type           coordinating   c
                 subordinating  s
- -------------- -------------- -
2 Formation      simple         s
                 compound       c
================================
```

**Table 10. Morphosyntactic specification of Number**
```
= ============== ============== =
P ATT            VAL            C
= ============== ============== =
1 Type           cardinal       c
                 ordinal        o
                 fractal        f
                 ordinal2       r
- -------------- -------------- -
4 Case           genitive       g
- -------------- -------------- -
6 Definiteness   no             n
                 yes            y
================================
```

Detailed description about attributes and the reasons for presenting such attributes can be found in [3]. Table 11 shows the number of possible meaningful tags for each PoS in Persian. There are 771 possible meaningful tags for Persian according to proposed MSD in MULTEXT-East. Verbs have the maximum number of tags among other PoSs.

**Table 11. The Number of Meaningful Tags for each PoS**

| Part of Speech | Number of Tags |
|---|---|
| Noun | 12 |
| Verb | 639 |
| Adjective | 12 |
| Pronoun | 78 |
| Determiner | 6 |
| Adverb | 3 |
| Adposition | 3 |
| Conjunction | 4 |
| Numeral | 12 |
| Interjection | 1 |
| Abbreviation | 1 |
| Total Number | 771 |

## 4- The Corpus

As mentioned before, in MULTEXT-East framework, the Orwell's *1984* is the main text and the corpus. So, the Persian version of *1984* has been annotated to prepare this resource for Persian. The corpus is comprised of:

- 110000 Tokens
- 12666 Paragraphs
- 6606 Sentences
- 6632 Lemmas
- 13597 Types

Among 771 possible meaningful tags for Persian, only 448 tags appeared in the corpus. The corpus is available on-line as part of MULTEXT-East package.

## 5- Conclusion

This article introduced the first steps toward preparing a Persian BLARK. In the first step toward this goal, we introduced an e-orthography for Persian and morphosyntactic specifications of Persian according to EAGLE/MULTEXT guidelines. This also comprises of introducing new PoS categorization of Persian and their MSDs in a way that it shows all the linguistic features of Persian in compatible with other languages in MULTEXT-East framework.

In the next step, the Persian version of Orwell's *1984* corpus is annotated. Having this corpus for Persian in the framework, a standard data-set is prepared for researchers who are interested in computational analysis of Persian. Moreover, this multilingual corpus enables researcher to investigate researches on Machine Translation applications especially the statistical approaches. In addition, introducing Persian to MULTEXT-East framework, allow researcher to study Persian in comparison with other languages in this framework.

As a future work, we have decided to expand the corpus to 1 million words. Moreover, a system that produces standard Persian e-text according to our proposed e-orthography for Persian e-text is under preparation.

## Acknowledgment

The authors would like to express their sincere gratitude to Prof. Tomaž Erjavec for many fruitful discussions. Also they would like to thank Prof. Damir Ćavar for his constructive idea.